\documentclass[11pt]{article}

\usepackage[preprint]{acl}

\usepackage{times}
\usepackage{latexsym}
\usepackage[T1]{fontenc}
\usepackage[utf8]{inputenc}
\usepackage{microtype}
\usepackage{inconsolata}

\usepackage{amsmath}
\usepackage{amssymb}
\usepackage{amsthm}

\usepackage{graphicx}
\usepackage{booktabs}
\usepackage{multirow}
\usepackage{makecell}
\usepackage{adjustbox}
\usepackage{tabularx}
\usepackage{array}
\usepackage[table]{xcolor}
\usepackage{colortbl}
\usepackage{capt-of}  

\usepackage{algorithm}
\usepackage{algorithmic}

\usepackage{tikz}
\usepackage{pgfplots}
\pgfplotsset{compat=1.18}
\usepgfplotslibrary{groupplots}
\usepgfplotslibrary{fillbetween}
\usetikzlibrary{intersections}
\usetikzlibrary{arrows.meta, positioning, fit, shapes.geometric, arrows}
\usepackage{xspace}

\definecolor{lightblue}{RGB}{135,206,250}
\definecolor{myblue}{RGB}{30,118,179}
\definecolor{mygreen}{RGB}{43,159,43}
\definecolor{myred}{RGB}{213,38,39}
\definecolor{myorange}{RGB}{255,126,13}
\definecolor{mycolor1}{RGB}{252, 132, 79}
\definecolor{mycolor2}{RGB}{145, 203, 118}
\definecolor{mycolor3}{RGB}{251, 200, 93}
\definecolor{mycolor4}{RGB}{237, 103, 104}
\definecolor{mycolor5}{RGB}{116, 191, 223}
\definecolor{mycolor6}{RGB}{60, 164, 115}
\definecolor{mycolor7}{RGB}{104, 212, 223}
\definecolor{mycolor8}{RGB}{155, 96, 180}
\definecolor{mycolor9}{RGB}{254, 186, 187}
\definecolor{mycolor10}{RGB}{119, 130, 245}
\definecolor{mycolor11}{RGB}{102, 177, 234}
\definecolor{mycolor12}{RGB}{180, 145, 138}

\usepackage{cleveref}

\title{One Size Does Not Fit All: Distribution-Guided Tensor-Wise \\ Adaptive Sparsification for Model Merging}

\author{
    \textbf{Yingfeng Luo\textsuperscript{1}\thanks{\xspace Equal contribution.}},
    \textbf{Dingyang Lin\textsuperscript{1}\footnotemark[1]},
    \textbf{Junxin Wang\textsuperscript{1}},
    \textbf{Kaiyan Chang\textsuperscript{1}},
    \textbf{Ziqiang Xu\textsuperscript{1}}, 
    \textbf{Weiqiao Shan\textsuperscript{1}} \\
    \textbf{Yinqiao Li\textsuperscript{2}}, 
    \textbf{Tong Zheng\textsuperscript{3}}, 
    \textbf{Bei Li}, 
    \textbf{Anxiang Ma\textsuperscript{1,4}},
    \textbf{Tong Xiao\textsuperscript{1,4}\thanks{\xspace Corresponding author.}},
    \textbf{Zhengtao Yu\textsuperscript{5}},
    \textbf{Jingbo Zhu\textsuperscript{1,4}}
    \\
    \textsuperscript{1} School of Computer Science and Engineering, Northeastern University, Shenyang, China\\
	\textsuperscript{2} City University of Hong Kong\ \
    \textsuperscript{3} University of Maryland, College Park  \\
    \textsuperscript{4} NiuTrans Research, Shenyang, China   \\
    \textsuperscript{5} Kunming University of Science and Technology, Kunming, China \\
    \texttt{luoyingfeng\_neu@outlook.com}; 
	\texttt{\{xiaotong,zhujingbo\}@mail.neu.edu.cn}
}

\begin{document}
\maketitle

\begin{abstract}
Merging multiple task-specific fine-tuned checkpoints into a single multi-task model provides an attractive alternative to joint multi-task training.
To mitigate cross-task interference, a common strategy is to sparsify task vectors before aggregation.
However, many widely used methods rely on a uniform global sparsity ratio, overlooking differences in pruning sensitivity across tasks and parameter tensors.
Consequently, such uniform sparsification may remove useful parameter updates from sensitive tensors while retaining redundant updates in others.
To address this limitation, we propose \textbf{TADrop} (\underline{\textbf{T}}ensor-wise \underline{\textbf{A}}daptive \underline{\textbf{Drop}}), a plug-and-play sparsifier that independently determines tensor-specific keep ratios from their magnitude distributions using a scale-invariant quantile ratio.
TADrop operates in a data-free manner by default, while offering optional lightweight module-level calibration when a small validation set is available.
Experiments across vision, language, and multimodal benchmarks, including LLM and LoRA merging, demonstrate consistent improvements across multiple host merging methods, including a 4.99-point gain over EMR-Merging on RoBERTa GLUE.
In many-task ViT-B/16 merging, the improvement increases with the number of tasks, reaching 2.26 points when merging 30 tasks.
Further analyses show that TADrop preserves more task-associated neurons under matched overall sparsity.
Together, these findings highlight the value of distribution-guided sparsification at the tensor level for improving model merging.

\end{abstract}

\section{Introduction}
Model merging provides a practical way to combine multiple task-specific fine-tuned models into a single multi-task model without joint retraining \citep{yang2024modelmergingllmsmllms,wang2025scaling}.
A widely adopted approach is based on \textit{task vectors} \citep{DBLP:conf/iclr/IlharcoRWSHF23}, where each fine-tuned checkpoint is represented as the sum of a shared pretrained model and a task-specific parameter update.
These task vectors are aggregated to construct a unified multi-task model.
However, directly aggregating these task vectors can lead to substantial cross-task interference.
A common strategy is therefore to sparsify task vectors before aggregation, removing low-magnitude updates that contribute little but can interfere when accumulated across tasks.

\begin{figure}[t!]
\centering
\includegraphics[width=0.99\linewidth]{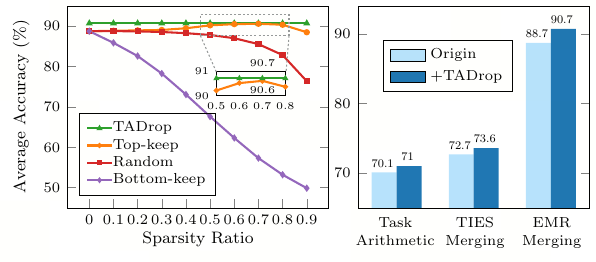}
\caption{\textbf{Left}: Average accuracy over eight ViT-B/32 tasks after EMR-Merging with different sparsification strategies and sparsity ratios.
\textbf{Right}: Average accuracy of different merging methods with and without TADrop.
}
\label{fig:tadrop_performance}
\end{figure}

\begin{figure*}[t!]
\centering
\includegraphics[width=0.99\linewidth]{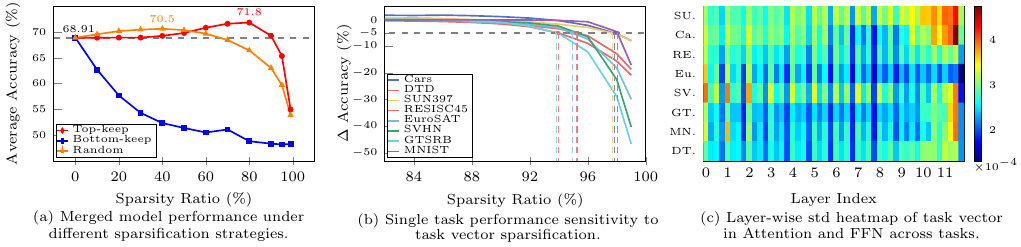}
\caption{Task vectors are heterogeneous across sparsity levels, tasks, and layers/modules, motivating tensor-wise adaptive sparsification. 
(a) illustrates how sparsification strategy and sparsity ratio affect merged-model performance; 
(b) shows that different tasks exhibit different sensitivity to task vector sparsification (Vertical markers denote 5-point accuracy drop thresholds); 
(c) visualizes the heterogeneity of task vector magnitude distributions across layers/modules and tasks.}
\label{fig:motivation}
\end{figure*}

Despite the effectiveness of sparsification, many existing methods apply a manually chosen global sparsity ratio uniformly across task vectors \citep{DBLP:conf/nips/YadavTCRB23,DBLP:conf/icml/WangDOFF24,DBLP:conf/iclr/YangW00G0T24}.
Such a uniform strategy implicitly assumes that different tasks and tensors can tolerate the same degree of sparsification.
In practice, however, their sensitivity to pruning can vary substantially.
As shown in \Cref{fig:motivation}a, merged performance is highly sensitive to the chosen global sparsity ratio, making it difficult to select reliably.
Moreover, different tasks exhibit varying sparsification tolerance (\Cref{fig:motivation}b), while magnitude distributions differ considerably across layers and modules (\Cref{fig:motivation}c).
Consequently, uniform sparsification can over-prune sensitive tensors while retaining potentially redundant updates in others.
These limitations become especially problematic in large-scale multi-task settings, where task heterogeneity increases and cross-task interference is amplified.

Recent studies have explored heterogeneity through layer-wise sparsity adaptation \citep{chopra2025lewis}, adaptive layer rescaling \citep{wang2026larv}, and structured component selection \citep{li2026cass}.
Despite these advances, conventional magnitude-based sparsification typically uses update magnitudes to determine which parameters to retain, but relies on a predefined global sparsity ratio to determine how many.
This raises a natural question: \emph{can tensor-wise magnitude distributions provide a useful signal for adapting sparsity levels across heterogeneous task-vector tensors?}
Such a tensor-local strategy would enable adaptive sparsification within individual task vectors before aggregation, complementing existing mechanisms for mitigating cross-task interference.

To this end, we propose \textbf{TADrop} (\underline{\textbf{T}}ensor-wise \underline{\textbf{A}}daptive \underline{\textbf{Drop}}), a lightweight sparsification method that independently determines tensor-specific keep ratios from their magnitude distributions.
Our key intuition is that the distribution of update magnitudes can inform how aggressively a tensor should be sparsified: tensors with a small number of dominant updates can be pruned more aggressively, whereas those with substantial updates spread across more entries may benefit from higher retention.
TADrop translates this intuition into a ratio of magnitude quantiles, enabling tensor-wise adaptive sparsification without requiring task data.
When a small validation set is available, optional module-level calibration can further refine the sparsification strength.
By processing task vectors before aggregation, TADrop can be integrated into existing merging methods without modifying their original aggregation rules.

We evaluate TADrop across three representative merging methods and diverse benchmarks spanning vision, language, and multimodal tasks, including large language models and LoRA-based merging.
TADrop consistently improves all three host merging methods on both ViT backbones (\Cref{fig:tadrop_performance}) and achieves a 4.99-point gain over EMR-Merging on RoBERTa GLUE.
Its benefits also become more pronounced as the number of merged tasks increases, reaching a 2.26-point improvement over EMR-Merging when merging 30 ViT-B/16 tasks.
Further analyses show that TADrop outperforms uniform sparsification under matched overall retention budgets and preserves more task-associated neurons, providing additional evidence for the benefits of tensor-wise adaptive sparsification.

Overall, this work makes three contributions.
First, we empirically examine task-vector sparsification across tasks and tensors, motivating the use of tensor-local magnitude distributions to guide adaptive pruning.
Second, we introduce TADrop, a self-contained and method-agnostic sparsifier that derives tensor-specific keep ratios from magnitude distributions, with optional lightweight calibration.
Third, we demonstrate consistent improvements across multiple merging methods and diverse benchmarks.
Taken together, our findings highlight the effectiveness of distribution-guided tensor-wise sparsity allocation as a complementary strategy for improving model merging.

\begin{figure*}[htbp]
\centering
\includegraphics[width=0.99\linewidth]{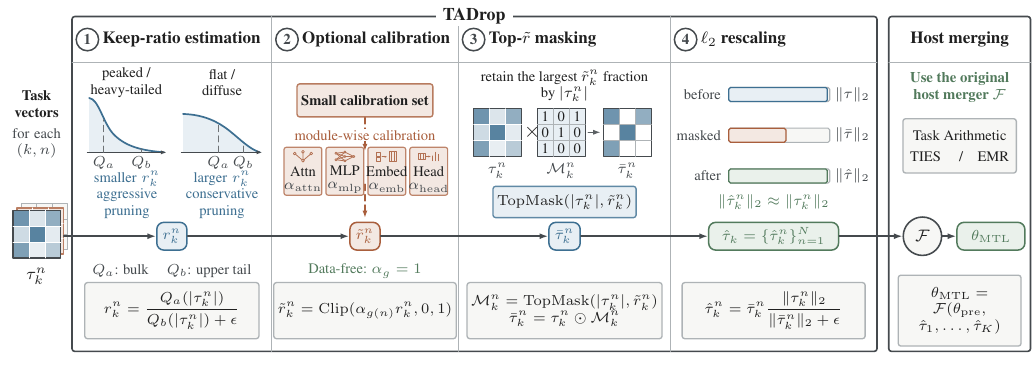}
\caption{\textbf{Overview of TADrop.} Given task vectors $\{\tau_k\}_{k=1}^{K}$, TADrop independently sparsifies each parameter tensor $\tau_k^n$ before host merging.  It (1) estimates a tensor-wise keep ratio $r_k^n$ from the magnitude distribution via a quantile ratio, (2) optionally calibrates the ratio by functional module group using a small calibration set, with $\alpha_g=1$ in the data-free setting, (3) retains the largest-magnitude $\tilde{r}_k^n$ fraction, and (4) rescales the masked tensor to approximately preserve $\lVert\tau_k^n\rVert_2$.
  The resulting sparsified task vectors $\{\hat{\tau}_k\}_{k=1}^{K}$ are passed to the original host merger $\mathcal{F}$ without modifying its merging rule.}
\label{fig:method}
\end{figure*}

\section{Related Work}

Task-vector merging often suffers from cross-task interference, motivating various strategies for filtering or reconciling conflicting parameter updates.
Task Arithmetic \citep{DBLP:conf/iclr/IlharcoRWSHF23} provides a simple foundation for combining task vectors.
Building upon this paradigm, TIES-Merging \citep{DBLP:conf/nips/YadavTCRB23} removes low-magnitude updates and resolves sign conflicts, while DARE \citep{DBLP:conf/icml/Yu0Y0L24} employs random dropping with rescaling.
EMR-Merging \citep{huang2024emr} further combines shared parameter updates with task-specific modulation.

Beyond conventional sparsification, recent studies have explored adaptive parameter selection and structure-aware merging to account for heterogeneity in task updates.
TALL-masks \citep{DBLP:conf/icml/WangDOFF24} and Localize-and-Stitch \citep{he2024localize} identify task-specific parameter subsets to preserve essential task information.
CABS \citep{DBLP:journals/corr/abs-2503-01874} combines conflict-aware masking with balanced sparsification to mitigate cross-task conflicts and layer-wise imbalance.
LEWIS \citep{chopra2025lewis} adapts layer-wise sparsity using activation-based importance estimated from calibration data, while LARV \citep{wang2026larv} introduces data-free layer-wise rescaling to account for heterogeneous layer characteristics.
At a finer granularity, NeuronMerge \citep{gu2025neuronmerge} groups MLP neurons by functionality, whereas CASS \citep{li2026cass} leverages attention heads and FFN neurons for structured sparsification.
Subspace-based methods, including ImPart \citep{DBLP:conf/acl/YangL0WY0C25} and AdaRank \citep{lee2026adarank}, exploit singular components and their importance to guide sparsification or component selection.
For parameter-efficient fine-tuning, KnOTS \citep{stoica2025model} aligns LoRA updates through SVD before applying existing merging methods.

TADrop takes a complementary approach by determining tensor-specific keep ratios directly from the magnitude distribution of each task-vector tensor.
These decisions are made independently for each task vector, allowing potentially redundant updates to be removed before aggregation without explicitly modeling cross-task conflicts.
TADrop operates in a data-free manner by default, with optional module-level calibration when a small validation set is available.
This self-contained design enables TADrop to complement existing merging methods without modifying their original aggregation rules.
A broader comparison of sparsification strategies is provided in Appendix~\Cref{sec:spar_works}.

\section{Methodology}
\label{sec:methodology}

\subsection{Preliminaries}

Let $\theta_{\mathrm{pre}} = \{\theta_{\mathrm{pre}}^{n}\}_{n=1}^{N}$ denote a pretrained model consisting of $N$ parameter tensors.
Starting from the same initialization, we independently fine-tune it on $K$ downstream tasks, obtaining task-specific checkpoints $\{\theta_k\}_{k=1}^{K}$.
For each task $k$, we define its task vector as
\begin{equation}
    \tau_k = \theta_k - \theta_{\mathrm{pre}},
\end{equation}
where $\tau_k = \{\tau_k^{n}\}_{n=1}^{N}$ and
$\tau_k^{n} = \theta_k^{n} - \theta_{\mathrm{pre}}^{n}$ denotes the update of the $n$-th parameter tensor.

We denote a generic task-vector merging method as
\begin{equation}
    \theta_{\mathrm{MTL}}
    =
    \mathcal{F}\left(
        \theta_{\mathrm{pre}},
        \tau_1,\ldots,\tau_K
    \right),
\label{eq:host_merging}
\end{equation}
where $\mathcal{F}$ denotes the host merging procedure, such as Task Arithmetic \citep{DBLP:conf/iclr/IlharcoRWSHF23}, TIES-Merging \citep{DBLP:conf/nips/YadavTCRB23}, or EMR-Merging \citep{huang2024emr}.
TADrop independently sparsifies each task vector before the resulting task vectors are passed to $\mathcal{F}$, while leaving the remaining merging procedure unchanged.

\subsection{Tensor-wise Adaptive Drop}

\paragraph{Distribution-guided keep ratio.}
Given a task-vector tensor $\tau_k^n$, TADrop estimates how aggressively it should be sparsified directly from its magnitude distribution.
The key intuition is that the relative shape of this distribution provides a useful proxy for how concentrated the task-specific updates are.
When large-magnitude updates are concentrated in a small fraction of entries, the tensor can be sparsified more aggressively.
Conversely, when they are distributed across a larger fraction of entries, more parameters should be retained.

We quantify this property using the ratio between two magnitude quantiles:
\begin{equation}
    r_k^n
    =
    \frac{
        Q_a\left(\left|\tau_k^n\right|\right)
    }{
        Q_b\left(\left|\tau_k^n\right|\right) + \epsilon
    },
    \qquad 0 < a < b < 1,
\end{equation}
where $Q_p(\cdot)$ denotes the $p$-th quantile and $\epsilon$ is a small constant for numerical stability.
We use $r_k^n$ as the tensor-specific keep ratio.
A smaller ratio indicates a sharper separation between the bulk of the entries and the upper tail, leading to more aggressive pruning, whereas a larger ratio indicates a more broadly distributed update and therefore a more conservative keep ratio.

The quantile ratio captures the relative contrast between typical and large-magnitude updates within each tensor, while reducing sensitivity to differences in their overall magnitude scales.
We use this statistic as a simple distribution-based signal to guide tensor-specific sparsification.
A formal justification is provided in Appendix~\Cref{sec:theory}.

\paragraph{Optional module-level calibration.}
The distribution-guided keep ratios $r_k^n$ can be directly applied to sparsify task vectors without requiring task data.
In practical scenarios where a small validation set is available, we additionally provide a lightweight calibration step to further refine the sparsification strength across functional modules.
Specifically, we partition parameter tensors into a small set of module groups
$\mathcal{G}=\{g_1,\ldots,g_M\}$,
such as attention projections, MLP projections, embeddings, and output heads.
For each group $g\in\mathcal{G}$, we introduce a scalar calibration factor $\alpha_g$ and define the calibrated keep ratio as
\begin{equation}
    \tilde{r}_k^n
    =
    \mathrm{Clip}\left(
        \alpha_{g(n)} r_k^n,
        0,
        1
    \right),
\end{equation}
where $g(n)$ denotes the module group containing the $n$-th tensor and $\mathrm{Clip}(\cdot,0,1)$ ensures a valid keep ratio.
We perform calibration at the functional module-group level, allowing tensors with similar functional roles to share a common adjustment.
The calibration factor $\alpha_g$ modulates the overall sparsification strength of each group while preserving the tensor-wise variation captured by $r_k^n$.
In the default data-free setting, we set $\alpha_g=1$ for all $g\in\mathcal{G}$, so that $\tilde{r}_k^n=r_k^n$.

\paragraph{Masking and norm-preserving rescaling.}
Given the final keep ratio $\tilde{r}_k^n$, we construct a binary mask that retains the largest $\tilde{r}_k^n$ fraction of entries in $\tau_k^n$ by magnitude:
\begin{equation}
    \mathcal{M}_k^n
    =
    \operatorname{TopMask}
    \left(
        \left|\tau_k^n\right|,
        \tilde{r}_k^n
    \right),
\end{equation}
where $\operatorname{TopMask}(\cdot,r)$ returns a binary mask that retains the largest $r$ fraction of entries by magnitude.
The masked tensor is then given by
\begin{equation}
    \bar{\tau}_k^n
    =
    \tau_k^n \odot \mathcal{M}_k^n,
\end{equation}
where $\odot$ denotes element-wise multiplication.

Since different tensors are sparsified to different degrees, masking can reduce their magnitudes by different amounts.
We therefore rescale each masked tensor to preserve its pre-sparsification $\ell_2$ norm:
\begin{equation}
    \hat{\tau}_k^n
    =
    \bar{\tau}_k^n
    \cdot
    \frac{
        \left\|\tau_k^n\right\|_2
    }{
        \left\|\bar{\tau}_k^n\right\|_2 + \epsilon
    }.
\end{equation}
The resulting sparsified task vector is
$\hat{\tau}_k=\{\hat{\tau}_k^n\}_{n=1}^{N}$.
TADrop then replaces each original task vector $\tau_k$ in \Cref{eq:host_merging} with its sparsified counterpart $\hat{\tau}_k$, while leaving the host merging procedure $\mathcal{F}$ unchanged.

\section{Experiments}

\begin{table*}[t!]
\centering
\resizebox{0.99\textwidth}{!}{\begin{tabular}{lccccccccc}
\toprule
\textbf{Methods} & \textbf{SUN397} & \textbf{Cars} & \textbf{RESISC45} & \textbf{EuroSAT} & \textbf{SVHN} & \textbf{GTSRB} & \textbf{MNIST} & \textbf{DTD} & \textbf{Avg.} \\
\midrule
{Individual} & 79.2 / 84.9 & 77.7 / 92.4 & 96.1 / 97.4 & 99.7 / 99.7 & 97.5 / 98.1 & {98.7} / 99.2 & 99.7 / 99.8 & 79.4 / {84.1} & 91.0 / 94.4 \\
{Traditional MTL} & {73.9} / {80.8} & {74.4} / {90.6} & {93.9} / {96.3} & {98.2} / {96.3} & {95.8} / {97.6} & 98.9 / {99.1} & {99.5} / {99.6} & {77.9} / 84.4 & {88.9} / {93.5} \\
\midrule
{Weight Averaging} & 65.3 / 72.1 & 63.4 / 81.6 & 71.4 / 82.6 & 71.7 / 91.9 & 64.2 / 78.2 & 52.8 / 70.7 & 87.5 / 97.1 & 50.1 / 62.8 & 65.8 / 79.6 \\
{Fisher Merging} & {68.6} / 69.2 & {69.2} / 88.6 & 70.7 / 87.5 & 66.4 / 93.5 & 72.9 / 80.6 & 51.1 / 74.8 & 87.9 / 93.3 & 59.9 / 70.0 & 68.3 / 82.2 \\
{RegMean} & 65.3 / 73.3 & 63.5 / 81.8 & 75.6 / 86.1 & 78.6 / 97.0 & 78.1 / 88.0 & 67.4 / 84.2 & 93.7 / 98.5 & 52.0 / 60.8 & 71.8 / 83.7 \\
PCB-Merging & 66.7 / 76.8 & 65.5 /  86.2 & 78.5 / 89.4 & 79.3 / 96.5 & 86.4 / 88.3 & 77.1 / 91 & 98.2 / 98.6 & 59.1 / 73.6 & 76.3 / 87.5 \\
Localize-and-Stitch & 67.2 / - & 68.3 / - & 81.8 / - & 89.4 / - & 87.9 / - & 86.6 / - & 94.8 / - & {62.9} / - & 79.9 / - \\
{AdaMerging++} & 66.6 / 79.4 & 68.3 / 90.3 & {82.2} / {91.6} & {94.2} / {97.4} & {89.6} / {93.4} & {89.0} / {97.5} & {98.3} / {99.0} & 60.6 / 79.2 & {81.1} / {91.0} \\
{ProDistill} & 68.9 / {77.7} & 71.2 / {90.0} & 89.9 / 94.4 & 99.4 / 99.5 & 96.1 / 97.7 & 95.3 / 98.3 & 99.5 / 99.6 & 68.0 / {78.2} & 86.0 / 91.9 \\
\midrule
{Task Arithmetic} & 55.1 / 74.1 & 54.9 / 82.1 & 66.7 / 86.7 & 77.2 / 93.8 & \underline{80.2} / 87.9 & \textbf{69.7} / 86.8 & \textbf{97.3} / 98.9 & 50.1  / 65.6 & 68.9 / 84.5 \\
 \hspace{10pt} + ImPart & 62.2 / 71.2 & 60.7 / 81.3 & 71.1 / 84.7 & 76.6 / 93 & 73.5 / 83.5 & 63.4 / 79.8 & 93.5 / 98.3 & 50.9 / 64.3 & 69.0 / 82.0 \\
  \hspace{10pt} + TALL-masks & \textbf{65.9} / \underline{74.5} & \textbf{64.1} / \underline{83.9} & \textbf{75.5} / \underline{88.8} & \textbf{80.4} / \textbf{95.6} & 78.4 / \underline{88.4} & \underline{69.6} / 85.0 & 95.9 / 98.9 & \textbf{55.1} / \underline{67.3} & \textbf{73.1} / \underline{85.3} \\
\rowcolor{gray!12} \hspace{10pt} + TADrop(w/o) & 62.7 / 74.3 & 61.7 / 82.9 & 72.2 / 87.3 & \underline{78.2} / 93.5 & 77.4 / 88.1 & 67.5 / \underline{87.7} & 95.9 / \underline{99.0} & 52.2 / 66.8 & 71.0 / 84.9 \\
\rowcolor{gray!20} \hspace{10pt} + TADrop & \underline{63.6} / \textbf{75.0} & \underline{62.5} / \textbf{85.5} & \underline{73.5} / \textbf{89.0} & 77.9 / \underline{94.5} & \textbf{80.7} / \textbf{91.0} & 68.8 / \textbf{90.5} & \underline{96.7} / \textbf{99.2} & \underline{52.8} / \textbf{68.5} & \underline{72.1} / \textbf{86.7} \\
\midrule
{TIES-Merging} & \underline{64.9} / 74.4 & \underline{64.3} / \underline{84.5} & 74.6 / 88.0 & 76.4 / 94.0 & 81.2 / 85.7 & 69.3 / 82.1 & 96.5 / 98.7 & 54.2 / \underline{67.7} & 72.7 / 84.4 \\
 \hspace{10pt} + ImPart & 60.7 / 73.6 & 59.3 / 82.8 & 70.6 / 87.1 & \underline{81.4} / 93.9 & 81.2 / 87.9 & \underline{71.8} / 86.5 & \underline{97.2} / 98.9 & 52.8 / 66.3 & 71.9 / 84.6 \\
 \hspace{10pt} + TALL-masks & \textbf{66.8} / 74.4 & \textbf{65.8} / \underline{84.5} & \textbf{75.6} / 87.7 & 77.6 / 94.1 & 80.6 / 85.2 & 69.7 / 81.5 & 96.2 / 98.6 & \textbf{55.3} / \underline{67.7} & 73.5 / 84.2 \\
\rowcolor{gray!12}  \hspace{10pt} + TADrop(w/o) & 64.8 / \textbf{76.0} & 63.0 / 83.6 & 74.1 / \underline{88.5} & \textbf{82.1} / \underline{94.3} & \underline{81.5} / \underline{89.9} & 71.1 / \underline{86.6} & 97.1 / \underline{99.0} & \underline{55.2} / 67.5 & \underline{73.6} / \underline{85.7} \\
\rowcolor{gray!20} \hspace{10pt} + TADrop & 64.4 / \underline{75.8} & 63.4 / \textbf{85.2} & \underline{75.2} / \textbf{89.1} & 80.7 / \textbf{94.9} & \textbf{85.5} / \textbf{91.3} & \textbf{74.4} / \textbf{90.2} & \textbf{97.9} / \textbf{99.2} & 54.4 / \textbf{68.6} & \textbf{74.5} / \textbf{86.8} \\
\midrule
{EMR-Merging} & \underline{75.2} / 83.2 & 72.8 / 90.7 & 93.5 / 96.8 & \textbf{99.5} / \textbf{99.7} & \underline{96.9} / 97.9 & 98.1 / \underline{99.1} & \underline{99.6} / \underline{99.7} & 74.4 / 82.7 & 88.7 / 93.7 \\
\rowcolor{gray!12} \hspace{10pt} + TADrop(w/o) & \textbf{78.7} / \underline{84.3} & \underline{78.6} / \underline{92.1} & \underline{95.2} / \textbf{97.4} & \underline{99.4} / \textbf{99.7} & \textbf{97.4} / \underline{98.1} & \underline{98.7} / \underline{99.1} & \textbf{99.7} / \textbf{99.8} & \underline{77.9} / \underline{83.5} & \underline{90.7} / \underline{94.2} \\
\rowcolor{gray!20} \hspace{10pt} + TADrop & \textbf{78.7} / \textbf{84.4} & \textbf{79.0} / \textbf{92.2} & \textbf{95.4} / \underline{97.2} & \textbf{99.5} / \textbf{99.7} & \textbf{97.4} / \textbf{98.2} & \textbf{98.8} / \textbf{99.2} & \underline{99.6} / \textbf{99.8} & \textbf{78.2} / \textbf{84.0} & \textbf{90.8} / \textbf{94.4} \\
\bottomrule
\end{tabular}}
\caption{Multi-task performance when merging ViT models on eight tasks (a / b: a = ViT-B/32, b = ViT-L/14). \textbf{Bold} numbers indicate the best score within each backbone block, and \underline{underlined} numbers indicate the second-best score.}
\label{tab:performance_vit}
\end{table*}

\subsection{Setup}
\paragraph{Models and datasets.}
We evaluate TADrop across vision, language, and multimodal settings.
For \textbf{vision}, we use CLIP-based ViT-B/32 and ViT-L/14 encoders \citep{DBLP:conf/icml/RadfordKHRGASAM21} on eight image classification tasks.
We additionally evaluate LoRA merging using rank-16 adapters from the KnOTS \citep{stoica2025model}.
For \textbf{language}, we evaluate both encoder-based and autoregressive models.
We use RoBERTa \citep{liu2019roberta} on GLUE tasks and merge three models fine-tuned from Qwen2.5-32B-Instruct and evaluated on GSM8K, ARC-Challenge, and MMLU-CS.
Additional Llama-2-7B experiments are provided in Appendix \Cref{sec:llama2}.
For \textbf{multimodal}, we use BEiT3-base \citep{DBLP:conf/cvpr/WangBDBPLAMSSW23} on five vision-language tasks, including generative tasks.

\paragraph{Baselines.}
We integrate TADrop into three representative merging methods: Task Arithmetic \citep{DBLP:conf/iclr/IlharcoRWSHF23}, TIES-Merging \citep{DBLP:conf/nips/YadavTCRB23}, and EMR-Merging \citep{huang2024emr}, covering linear aggregation, magnitude-based sparsification, and task-specific modulation.
We also compare against TALL-masks \citep{DBLP:conf/icml/WangDOFF24}, subspace-based ImPart \citep{DBLP:conf/acl/YangL0WY0C25}, and Weight Averaging, Fisher Merging, and RegMean baselines.
For LoRA merging, we additionally consider KnOTS-TIES \citep{stoica2025model}.

\paragraph{Implementation details.}
We use $(a,b)=(0.50,0.95)$ unless otherwise specified.
We report two variants: TADrop(w/o), the data-free variant with $\alpha_g=1$ for all module groups, and TADrop, which additionally calibrates the module-level factors $\{\alpha_g\}$ using Optuna \citep{akiba2019optuna} with 32 validation samples per task.
In Task Arithmetic, TADrop sparsifies task vectors before aggregation.
In TIES-Merging, it replaces magnitude-based pruning while preserving sign-conflict resolution.
In EMR-Merging, its adaptive mask is combined with the consensus sign mask through element-wise multiplication.

\subsection{Main Results}

\paragraph{Performance on ViT models.}
\Cref{tab:performance_vit} presents the results of merging ViT-B/32 and ViT-L/14 models across eight image classification tasks.
TADrop consistently improves the average accuracy of Task Arithmetic, TIES-Merging, and EMR-Merging on both backbones.
For the strong EMR-Merging baseline, TADrop improves the average accuracy by 2.1 and 0.7 points, reaching 90.8 and 94.4 on ViT-B/32 and ViT-L/14, respectively, approaching the performance of individually fine-tuned models.
The data-free TADrop(w/o) also consistently improves all three merging methods, indicating that the distribution-guided keep ratios are effective without calibration.
Compared with existing sparsification methods, TADrop outperforms ImPart on both backbones and generally surpasses TALL-masks, except when combined with Task Arithmetic on ViT-B/32.

\begin{table*}[htp]
\centering
\resizebox{\textwidth}{!}{
\begin{tabular}{lccccccccc}
\toprule
\multirow{2}{*}{\textbf{Methods}} & \textbf{COCO-Re} & \multicolumn{4}{c}{\textbf{COCO-Captioning}} & \textbf{ImageNet} & \textbf{NLVR2} & \textbf{VQAv2} & \textbf{Avg.} \\
\cmidrule(lr){2-2} \cmidrule(lr){3-6} \cmidrule(lr){7-7} \cmidrule(lr){8-8} \cmidrule(lr){9-9} \cmidrule(lr){10-10}
 & Acc & BLEU & CIDEr$\times100$ & METEOR & ROUGE & Acc & Acc & Acc & Score \\
\midrule
Individual & 84.6 & 39.4 & 134.0 & 31.1 & 60.1 & 85.4 & 77.7 & 84.4 & 74.6 \\
\midrule
Weight Averaging & 18.9 & 3.1 & 0.1 & 11.5 & 15.9 & 67.7 & 28.0 & 62.9 & 26.0 \\
Task Arithmetic~\citep{DBLP:conf/iclr/IlharcoRWSHF23} & 31.8 & 3.3 & 0.0 & 11.8 & 17.6 & 70.8 & 38.1 & 69.3 & 30.3 \\
TIES-Merging~\citep{DBLP:conf/nips/YadavTCRB23} & 39.3 & 2.9 & 0.1 & 10.8 & 16.7 & 69.8 & 32.1 & 67.2 & 29.9 \\
\midrule
EMR-Merging \citep{huang2024emr} & 79.5 & 28.9 & 106.0 & \underline{27.2} & 53.4 & 77.4 & 74.8 & \underline{72.1 } & 64.9 \\
\rowcolor{gray!12} \hspace{10pt} + TADrop(w/o) & \underline{80.5} & \underline{32.4} & \textbf{119.0} & \textbf{28.7} & \underline{56.0} & \underline{80.1} & \underline{77.3} & \underline{72.1} & \underline{68.3} \\
\rowcolor{gray!20} \hspace{10pt} + TADrop & \textbf{80.9} & \textbf{33.7} & \underline{118.0} & \textbf{28.7} & \textbf{56.2} & \textbf{80.5} & \textbf{77.4} & \textbf{72.4} & \textbf{68.5} \\
\bottomrule
\end{tabular}
}
\caption{Performance of merging multimodal BEiT3 models on five vision-language tasks.}
\label{tab:performance_beit} 
\end{table*}

\paragraph{Performance on RoBERTa models.}
\Cref{tab:roberta_glue} reports the average performance of merging RoBERTa-base models across eight GLUE tasks.
TADrop consistently improves all three merging methods, achieving gains of $3.13$, $5.60$, and $4.99$ points over Task Arithmetic, TIES-Merging, and EMR-Merging, respectively.
Both the data-free and calibrated variants outperform ImPart and TALL-masks when integrated with Task Arithmetic or TIES-Merging.
Detailed results for individual GLUE tasks are provided in Appendix~\Cref{tab:roberta_glue_aligned}.

\begin{table}[t!]
\centering
\resizebox{0.48\textwidth}{!}{
\begin{tabular}{lc}
\toprule
\textbf{Method} & \textbf{Avg.} \\
\midrule
Individual & 85.56 \\
\midrule
Weight Averaging & 51.34 \\
Fisher Merging~\citep{DBLP:conf/nips/MatenaR22} & 45.57 \\
RegMean~\citep{DBLP:conf/iclr/Jin0P023} & 70.01 \\
\midrule
Task Arithmetic~\citep{DBLP:conf/iclr/IlharcoRWSHF23} & 66.65 \\
\hspace{10pt} + ImPart~\citep{DBLP:conf/acl/YangL0WY0C25} & 65.72 \\
\hspace{10pt} + TALL-masks~\citep{DBLP:conf/icml/WangDOFF24} & 54.55 \\
\rowcolor{gray!12} \hspace{10pt} + TADrop(w/o) & \underline{68.35} \\
\rowcolor{gray!20} \hspace{10pt} + TADrop & \textbf{69.78} \\
\midrule
TIES-Merging~\citep{DBLP:conf/nips/YadavTCRB23} & 64.04 \\
\hspace{10pt} + ImPart~\citep{DBLP:conf/acl/YangL0WY0C25} & 61.45 \\
\hspace{10pt} + TALL-masks~\citep{DBLP:conf/icml/WangDOFF24} & 66.45 \\
\rowcolor{gray!12} \hspace{10pt} + TADrop(w/o) & \underline{67.51} \\
\rowcolor{gray!20} \hspace{10pt} + TADrop & \textbf{69.64} \\
\midrule
EMR-Merging~\citep{huang2024emr} & 80.18 \\
\rowcolor{gray!12} \hspace{10pt} + TADrop(w/o) & \underline{84.94} \\
\rowcolor{gray!20} \hspace{10pt} + TADrop & \textbf{85.17} \\
\bottomrule
\end{tabular}
}
\caption{Average multi-task performance when merging RoBERTa-base models fine-tuned on eight GLUE text classification tasks. }
\label{tab:roberta_glue}
\end{table}

\begin{table}[t]
\centering
{\footnotesize
\setlength{\tabcolsep}{3pt}
\renewcommand{\arraystretch}{1.12}
\begin{tabularx}{\columnwidth}{
    @{}l*{4}{>{\centering\arraybackslash}X}@{}
}
\toprule
\textbf{Model} & \textbf{GSM8K} & \textbf{ARC-C} &
\mbox{\textbf{MMLU-CS}} & \textbf{Avg.} \\
\midrule
Qwen2.5-32B-Ins
 & 77.94 & 74.49 & 83.00 & 78.48 \\
\midrule
SAND-Math
 & 89.39 & 73.72 & 83.00 & 82.04 \\
OpenThinker
 & 60.65 & 72.27 & 86.00 & 72.97 \\
SWE-Swiss
 & 78.70 & 71.25 & 84.00 & 77.98 \\
\midrule
Task Arithmetic
 & 82.11 & 73.29 & \textbf{85.00} & 80.13 \\
\rowcolor{gray!12}
\hspace{5pt}+ TADrop(w/o)
 & \textbf{84.69} & \textbf{73.98} & 83.00 & \textbf{80.55} \\
\midrule
TIES-Merging
 & 83.70 & 73.72 & 84.00 & 80.47 \\
\rowcolor{gray!12}
\hspace{5pt}+ TADrop(w/o)
 & \textbf{83.85} & \textbf{74.15} & \textbf{84.00} & \textbf{80.67} \\
\midrule
EMR-Merging
 & 76.72 & 71.93 & \textbf{84.00} & 77.55 \\
\rowcolor{gray!12}
\hspace{5pt}+ TADrop(w/o)
 & \textbf{88.40} & \textbf{73.81} & 83.00 & \textbf{81.74} \\
\bottomrule
\end{tabularx}
}
\caption{Performance of merging three models fine-tuned from Qwen2.5-32B-Instruct on three benchmarks.}
\label{tab:llm_performance}
\end{table}

\begin{table}[t!]
\centering
\small
\renewcommand{\arraystretch}{1.1}
\begin{tabularx}{\columnwidth}{>{\raggedright\arraybackslash}Xr}
\toprule
\textbf{Method} & \textbf{Avg.} \\
\midrule
Task Arithmetic \citep{DBLP:conf/iclr/IlharcoRWSHF23}  & 61.35 \\
\rowcolor{gray!12}
\hspace{10pt} + TADrop(w/o) & \textbf{63.32} \\
\midrule
TIES-Merging \citep{DBLP:conf/nips/YadavTCRB23} & 63.62 \\
\rowcolor{gray!12}
\hspace{10pt} + TADrop(w/o) & \textbf{64.19} \\
\midrule
KnOTS-TIES \citep{stoica2025model} & 68.30 \\
\rowcolor{gray!12}
\hspace{10pt} + TADrop(w/o) & \textbf{69.16} \\
\bottomrule
\end{tabularx}
\caption{Average performance of LoRA model merging on the eight vision classification tasks using CLIP ViT-B/32 with rank-16 adapters.}
\label{tab:lora_merging}
\end{table}

\paragraph{Performance on LLMs.}
As shown in \Cref{tab:llm_performance}, TADrop(w/o) improves the average performance of Task Arithmetic, TIES-Merging, and EMR-Merging by 0.42, 0.20, and 4.19 points, respectively.
These results suggest that tensor-wise adaptive sparsification can also benefit large autoregressive language models across different merging methods.
Additional results on Llama-2 are provided in Appendix~\Cref{sec:llama2}.

\paragraph{Performance on BEiT3 models.}
\Cref{tab:performance_beit} presents results on five vision-language tasks using BEiT3-base, covering retrieval, captioning, classification, and visual reasoning.
TADrop improves the strong EMR-Merging baseline across all reported metrics, increasing the average score from $64.9$ to $68.5$.
These improvements across different task types further demonstrate the applicability of adaptive sparsification to heterogeneous multimodal merging.

\paragraph{Performance with LoRA fine-tuning.}
As shown in \Cref{tab:lora_merging}, TADrop(w/o) improves the average performance of Task Arithmetic, TIES-Merging, and KnOTS-TIES by 1.97, 0.57, and 0.86 points, respectively.
The improvement with KnOTS-TIES further suggests that TADrop can complement subspace-based adapter merging, extending its applicability beyond full-parameter fine-tuning.
Detailed results for individual tasks are provided in Appendix~\Cref{tab:lora_full}.

\begin{figure}[t!]
    \centering
    \resizebox{0.99\linewidth}{!}{
    \begin{tikzpicture}

\newcommand{\mybarwidth}{7.5pt}
    
\begin{groupplot}[
group style={
    group size=2 by 1,
    horizontal sep=1cm,
    group name=myplots
},
set layers, 
tick align=inside,
]
\nextgroupplot[
    at={(0,0)},
    anchor=south west,
    width=0.46\textwidth,
    height=0.30\textheight,
    grid=major,
    major grid style={dashed},
    enlarge x limits=0.10,
    nodes near coords,
    nodes near coords align={south},
    nodes near coords style={text=black, font=\small, /pgf/number format/fixed, /pgf/number format/fixed zerofill, /pgf/number format/precision=2},
    ymin=87, ymax=94,
    yticklabel style={/pgf/number format/fixed, /pgf/number format/fixed zerofill, /pgf/number format/precision=2},
    ylabel={Average Performance},
    legend entries={EMR-Merging, + TADrop, Individual},
    legend style={
        at={(axis cs:24.8,89.1)},
        anchor=north,
        legend cell align=left,
        font=\small,
    },
    xmin=8, xmax=30.5,
    xtick={8,16,24,30},
    xlabel={Number of Tasks},
]

\addplot[name path=EMR,lightblue, line width=1pt, mark=square*, mark size=1.5pt] coordinates {
    (8,87.56)(16,89.59)(24,90.85)(30,89.59)
};

\addplot[name path=TADrop,myblue, line width=1pt, mark=pentagon*, mark size=1.5pt] coordinates {
    (8,88.30)(16,90.57)(24,92.16)(30,91.85)
};

\addplot[mygreen, line width=1pt, mark=*, mark size=1.5pt] coordinates {
    (8,89.84)(16,91.90)(24,93.01)(30,93.02)
};

\addplot[fill=blue!30!gray!10] fill between [of=EMR and TADrop,];

\nextgroupplot[
    width=0.3\textwidth,
    height=0.3\textheight,
    ybar, 
    bar width=2.0*\mybarwidth, 
    ymin=0, ymax=2.5,
    title style={at={(0.5, -0.2)}, yshift=-17pt},
    xtick={0,1,2,3},
    xticklabels={8,16,24,30},
    xtick=data,
    xtick align=inside,
    xlabel={Number of Tasks},
    enlarge x limits=0.25,
    nodes near coords,
    nodes near coords style={scale=1, font=\small, xshift=-1pt, yshift=0pt},
    legend style={
        fill=none,
        column sep=0.em,
        font=\scriptsize, 
        legend cell align=left,
        at={(0.67, 0.83)}, 
        legend columns=1, 
    },
]
\addplot[draw=none, fill=myblue, bar shift=0*\mybarwidth, area legend] coordinates {
    (0,0.74) (1,0.98) (2,1.31) (3,2.26)
};
\end{groupplot}
\end{tikzpicture}
    }
    \caption{Scalability of TADrop in 30-task model merging on ViT-B/16. \textbf{Left}: Absolute average accuracy as the number of merged tasks grows from 8 to 30. \textbf{Right}: Performance gain over the EMR-Merging baseline.}
    \label{fig:many_tasks}
\end{figure}

\section{Analyses}

\subsection{Scalability in Many-Task Merging}

To examine how TADrop scales with the number of merged tasks, we progressively merge nested subsets of 8, 16, 24, and 30 ViT-B/16 tasks, using EMR-Merging as the baseline.
As shown in \Cref{fig:many_tasks}, TADrop consistently improves average accuracy across all evaluated task counts, with gains increasing from 0.74 points for 8 tasks to 2.26 points for 30 tasks.
When the number of tasks increases from 24 to 30, EMR-Merging drops by 1.26 points, whereas TADrop limits this decline to 0.31 points.
These results suggest that the benefits of adaptive sparsification become more pronounced as the number of merged tasks increases.
Detailed results for all 30 tasks are provided in Appendix~\Cref{tab:30_vit_full}.

\subsection{Sparsity Level and Tensor-wise Allocation}

\begin{figure}[t!]
    \centering
    \includegraphics[width=0.98\linewidth]{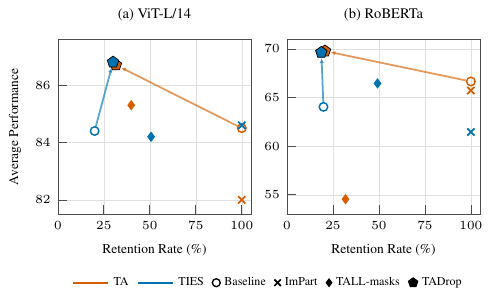}
    \caption{Average performance versus task-vector retention on ViT-L/14 and RoBERTa. Colors distinguish host merging methods, while markers indicate different sparsification strategies. }
    \label{fig:retention_performance}
\end{figure}

To understand whether the improvements of TADrop primarily arise from changes in overall sparsity or from its tensor-wise adaptive strategy, we examine the relationship between task-vector retention and merging performance.
Here, the retention ratio denotes the fraction of task-vector entries preserved after sparsification.
As shown in \Cref{fig:retention_performance}, TADrop consistently improves Task Arithmetic and TIES-Merging on both ViT-L/14 and RoBERTa, although the changes in overall retention vary across settings.
For example, when applied to TIES-Merging, TADrop increases the retention ratio from 20.00\% to 30.00\% on ViT-L/14, while reducing it from 19.86\% to 18.67\% on RoBERTa.
Despite these opposite changes in retention, average performance improves in both cases, from 84.40 to 86.80 and from 64.04 to 69.64, respectively.
These observations suggest that the improvements of TADrop are not simply associated with consistently increasing or decreasing the overall retention ratio.

To further investigate the importance of sparsity allocation, we compare TADrop against uniform Top-$k$ sparsification under matched overall retention budgets.
Unlike TADrop, which assigns tensor-specific keep ratios based on magnitude distributions, uniform Top-$k$ applies the same retention ratio across tensors.
As shown in \Cref{tab:matched_retention_control}, TADrop outperforms uniform Top-$k$ in all four settings, with improvements ranging from 1.33 to 4.06 points.
For instance, on RoBERTa with TIES-Merging, both methods retain 18.67\% of task-vector entries, yet TADrop achieves an average score of 69.64 compared with 65.58 for uniform Top-$k$.
Together, these results suggest that the effectiveness of task-vector sparsification depends not only on the overall retention budget, but also on how sparsity is allocated across tensors, supporting the value of distribution-guided tensor-wise adaptation.

\begin{table}[t]
\centering
{\scriptsize
\setlength{\tabcolsep}{2.0pt}
\renewcommand{\arraystretch}{1.08}
\begin{tabular*}{\columnwidth}{@{\extracolsep{\fill}}llccc@{}}
\toprule
\makecell{\textbf{Backbone}} & \makecell{\textbf{Method}} &
\makecell{\textbf{Baseline}} &
\makecell{\textbf{w/ Top-$k$}} &
\makecell{\textbf{w/ TADrop}} \\
\midrule
\multirow{2}{*}{ViT-L/14}
  & TA   & 100.00 / 84.50 & 31.54 / 84.34 & 31.54 / \textbf{86.70} \\
  & TIES & 20.00 / 84.40  & 30.00 / 84.64 & 30.00 / \textbf{86.80} \\
\midrule
\multirow{2}{*}{RoBERTa}
  & TA   & 100.00 / 66.65 & 20.65 / 68.45 & 20.65 / \textbf{69.78} \\
  & TIES & 19.86 / 64.04  & 18.67 / 65.58 & 18.67 / \textbf{69.64} \\
\bottomrule
\end{tabular*}
}
\caption{Comparison of uniform Top-$k$ and TADrop under matched retention budgets. Each entry reports retention (\%) / average performance. Baseline denotes the original host method.}
\label{tab:matched_retention_control}
\end{table}

\subsection{Task-Associated Neuron Analysis}

The preceding analysis shows that TADrop outperforms uniform Top-$k$ sparsification under matched overall retention budgets.
To investigate whether this advantage is associated with better preservation of task-relevant information, we examine task-associated neurons in the MLP expansion layers.
We construct a shared probe set by pooling samples from all tasks and assign each neuron a task label based on the majority task identity among its top-$k$ activating samples.
For example, if eight of a neuron's top-10 activating samples come from Cars and two from MNIST, the neuron is assigned the Cars label.
We then compare TADrop with uniform-ratio magnitude-based Top-$k$ sparsification, matching both the overall retention ratio and the tensor-wise masking granularity.

\begin{figure}[t!]
    \centering
    \includegraphics[width=0.98\linewidth]{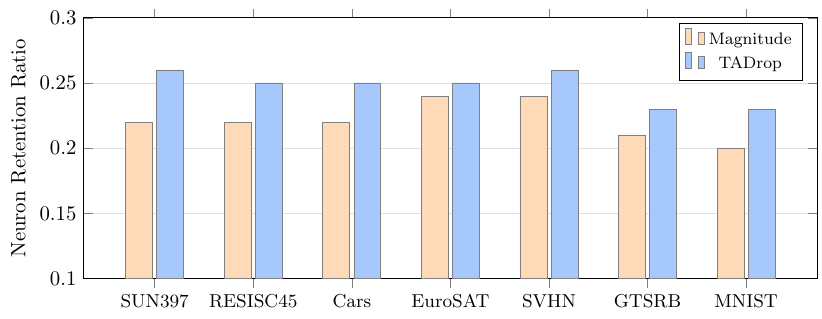}
    \caption{Retention of task-labeled neurons under matched overall sparsity. TADrop consistently outperforms uniform-ratio magnitude-based sparsification.}
    \label{fig:neuron_retention}
\end{figure}

As shown in \Cref{fig:neuron_retention}, TADrop retains a higher fraction of task-labeled neurons across all seven evaluated tasks under the same overall sparsity.
This suggests that distribution-guided sparsity allocation may better preserve task-associated neurons than uniform sparsification, offering a possible functional-level explanation for its improved merging performance.

\subsection{Ablation Study}

\begin{figure}[t!]
    \centering
    \includegraphics[width=0.99\linewidth]{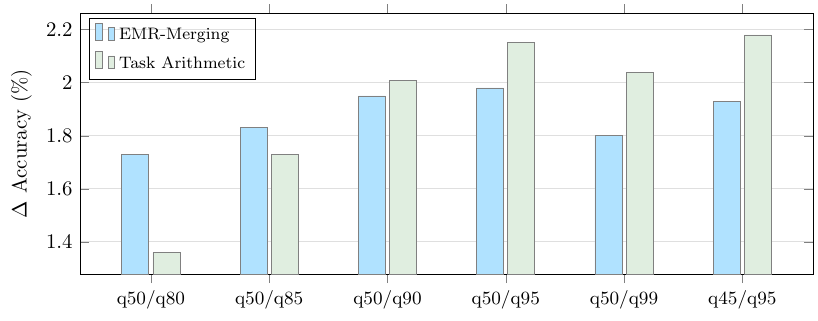}
    \caption{Ablation on quantile choices \((Q_a,Q_b)\) on ViT-B (8 tasks). Bars show the change in average accuracy relative to each host method's baseline.}
    \label{fig:quantile_ablation}
\end{figure}

\paragraph{Quantile selection.}
We examine the sensitivity of TADrop to different quantile pairs $(a,b)$ on eight ViT-B/32 tasks using EMR-Merging and Task Arithmetic.
As shown in \Cref{fig:quantile_ablation}, TADrop consistently improves both methods across all tested configurations, indicating that its effectiveness does not rely on carefully tuned quantile levels.
We adopt $(a,b)=(0.50,0.95)$ as the default due to its strong performance across both methods.

\paragraph{Rescaling strategy.}
We compare three rescaling strategies: no rescaling (w/o R), scaling by $1/(1-p)$ with drop rate $p$, and our tensor-wise $\ell_2$ norm preservation.
The $1/(1-p)$ correction is commonly used in random sparsification to preserve the expected magnitude of parameter updates \citep{DBLP:conf/icml/Yu0Y0L24}.
As shown in \Cref{tab:rescale_ablation}, $\ell_2$ rescaling achieves the best performance (90.8), outperforming both no rescaling (86.8) and $1/(1-p)$ correction (71.1).
Unlike random dropping, magnitude-based sparsification preferentially retains large updates, so the $1/(1-p)$ correction can over-amplify the retained parameters, particularly at high sparsity levels.
These results suggest that tensor-wise norm preservation provides a more suitable rescaling strategy for TADrop.

\begin{table}[t]
\centering
\setlength{\tabcolsep}{4pt}
\small
\begin{tabular}{lcccc}
\toprule
Method & EMR-Merging & w/o R & $(1-p)$ R & $\ell_2$ R \\
\midrule
Avg. & 88.7 & 86.8 & 71.1 & \textbf{90.8} \\
\bottomrule
\end{tabular}
\caption{Ablation results of rescaling strategies on the 8 ViT tasks.
R denotes the rescaling operation.}
\label{tab:rescale_ablation}
\end{table}

\paragraph{Effect of module-level calibration.}
To assess the contribution of module-level calibration, we compare TADrop with its data-free variant, TADrop(w/o), using the results in \Cref{tab:performance_vit,tab:performance_beit,tab:roberta_glue}.
The data-free variant already achieves substantial improvements over the host merging methods, indicating that the distribution-guided keep ratios are effective without calibration.
Module-level calibration provides further gains, improving EMR-Merging from 90.7 to 90.8 on ViT-B/32, from 84.94 to 85.17 on RoBERTa, and from 68.3 to 68.5 on BEiT3.
Larger improvements are observed in some settings, such as TIES-Merging on RoBERTa (67.51 to 69.64).
These results show that TADrop is effective in its data-free setting, while optional calibration can further improve performance when validation data are available.

\section{Conclusion}
In this paper, we introduced TADrop, a lightweight and method-agnostic approach that adapts task-vector sparsification to the magnitude distribution of each parameter tensor.
By assigning tensor-specific keep ratios before aggregation, TADrop can be integrated into existing merging methods without modifying their aggregation rules.
Experiments across vision, language, and multimodal benchmarks, including LLM and LoRA merging, demonstrate consistent improvements, with increasingly pronounced gains as more tasks are merged.
Further analyses show that TADrop outperforms uniform sparsification under matched overall retention budgets and preserves more task-associated neurons, supporting the benefits of adaptive sparsity allocation.
These findings suggest that improving model merging is not solely about how task vectors are aggregated, but also about how they are selectively sparsified before aggregation.

\section*{Limitations}
TADrop relies on tensor-wise magnitude distributions to guide sparsity allocation, providing a simple and lightweight alternative to more elaborate importance estimation.
Although effective across diverse merging settings, magnitude statistics provide an indirect signal for sparsification and may not fully capture the functional characteristics of different parameter tensors.
Future work could explore complementary signals for sparsity allocation while preserving the simplicity and compatibility of TADrop with existing merging methods.


\bibliography{custom}

\appendix
\clearpage

\begin{table*}[t!]
\centering
\footnotesize
\setlength{\tabcolsep}{4pt}
\renewcommand{\arraystretch}{1.10}
\begin{tabularx}{\textwidth}{@{}
>{\raggedright\arraybackslash}p{0.19\textwidth}
>{\raggedright\arraybackslash}p{0.24\textwidth}
>{\raggedright\arraybackslash}X
>{\raggedright\arraybackslash}p{0.16\textwidth}
@{}}
\toprule
\textbf{Method} &
\textbf{Sparsification Signal} &
\textbf{Sparsity Allocation / Selection} &
\textbf{Auxiliary Data} \\
\midrule
\textbf{TADrop (Ours)}
& Tensor magnitude quantiles
& Tensor-wise adaptive keep ratios with magnitude-based selection
& Optional \\
\midrule
TIES-Merging \citep{DBLP:conf/nips/YadavTCRB23}
& Update magnitudes
& Uniform keep ratio with magnitude-based selection
& None \\

DARE \citep{DBLP:conf/icml/Yu0Y0L24}
& Random sampling
& Uniform random drop probability
& None \\

DELLA-Merging \citep{DBLP:journals/corr/abs-2406-11617}
& Magnitude ranking
& Magnitude-dependent parameter dropping probabilities
& None \\
\midrule
TALL-masks \citep{DBLP:conf/icml/WangDOFF24}
& Task-specific update contrasts
& Task-specific element-wise masks
& Validation for tuning \\

PCB-Merging \citep{du2024parameter}
& Intra-task importance and inter-task similarity
& Element-wise pruning and rescaling
& None \\

CABS \citep{DBLP:journals/corr/abs-2503-01874}
& Cross-task conflicts and weight magnitudes
& Conflict-aware masks with structured $n$:$m$ sparsity
& None \\
\midrule
ImPart \citep{DBLP:conf/acl/YangL0WY0C25}
& Singular-value importance
& Adaptive sparsity ratios across singular vectors
& Validation for tuning \\

LEWIS \citep{chopra2025lewis}
& Activation-based layer importance
& Layer-wise adaptive sparsity
& Calibration data \\

AdaRank \citep{lee2026adarank}
& Prediction entropy
& Adaptive selection of singular components
& Unlabeled data \\

CASS \citep{li2026cass}
& Activation-based component contributions
& Structured masks over attention heads and FFN neurons
& Unlabeled data \\
\bottomrule
\end{tabularx}
\caption{
Conceptual comparison of representative task-vector sparsification and selection methods in terms of their decision signals, parameter retention strategies, and auxiliary data requirements.
}
\label{tab:merging_comparison}
\end{table*}

\section{Conceptual Comparison of Sparsification Methods}
\label{sec:spar_works}

As summarized in \Cref{tab:merging_comparison}, existing methods employ diverse signals and retention strategies for task-vector sparsification.
TADrop uses tensor-wise magnitude distributions to guide sparsity allocation, offering a simple and data-free approach to adaptive sparsification.
The comparison highlights the different design choices and data requirements of representative methods.

\section{Results on Llama-2}
\label{sec:llama2}
We further evaluate TADrop in the Llama-2-7B setting, using three task-specific models: Llama-2-7B-Chat \citep{DBLP:journals/corr/abs-2307-09288} for chat and knowledge tasks, MAmmoTH-7B \citep{yue2024mammoth} for mathematics, and Llama-2-Coder-7B \citep{manuel_romero_2023} for code generation.
We evaluate the merged models on six benchmarks: MMLU \citep{hendrycks2020measuring}, HellaSwag \citep{zellers2019hellaswag}, TruthfulQA \citep{lin2022truthfulqa}, GSM8K \citep{cobbe2021training}, HumanEval \citep{chen2021evaluating}, and MBPP \citep{austin2021program}.
Due to limited compute, we report results on a fixed subset of 100 test examples per benchmark, shared across all methods.

Table~\ref{tab:llama2_results} summarizes the results under this subset setting.
Across both methods, adding TADrop generally improves or matches the corresponding baseline on most benchmarks, with clearer gains on knowledge-style and math evaluations (e.g., MMLU and GSM8K), while maintaining comparable performance on code benchmarks (HumanEval/MBPP).

\section{Theoretical Properties of the Quantile Ratio}
\label{sec:theory}

This section examines two properties of the quantile ratio used in TADrop: scale invariance and sensitivity to distributional shape.
These properties provide theoretical motivation for using the ratio to guide tensor-wise sparsity allocation.

Empirical evidence suggests that parameter updates in neural networks often exhibit non-Gaussian, heavy-tailed behavior.
Motivated by prior analyses that model neural network parameters/updates with generalized Gaussian families\citep{DBLP:journals/corr/abs-2506-00486,wu2025backslashrateconstrainedoptimized}, we use the \textbf{Generalized Gaussian Distribution (GGD)} as a convenient parametric model for task-vector entries.

Let \(X\) denote a signed coefficient in a task vector tensor and assume \(X\sim\mathrm{GGD}(0,\alpha,\beta)\) with density
\begin{equation}
    f_X(x; \alpha, \beta) = \frac{\beta}{2\alpha \Gamma(1/\beta)} 
    \exp\!\left( - \left( \frac{|x|}{\alpha} \right)^\beta \right),
\end{equation}
where \(\alpha>0\) is a scale parameter and \(\beta>0\) controls peakedness and tail behavior.
We focus on the magnitude \(Y=|X|\), since TADrop operates on \(|\tau|\).

TADrop uses a quantile ratio \(R=Q_a/Q_b\) of the magnitude distribution to guide tensor-wise retention.
Let \(Q(p)\) denote the quantile function of \(Y=|X|\), i.e., \(\Pr(Y\le Q(p))=p\).
For the GGD, the CDF of \(Y\) satisfies
\begin{equation}
    \Pr(Y \le y) = P\!\left(\frac{1}{\beta}, \left(\frac{y}{\alpha}\right)^\beta\right),
\end{equation}
where \(P(\cdot,\cdot)\) is the regularized lower incomplete gamma function.
Solving for the quantile yields
\begin{equation}
    Q(p; \alpha, \beta) = \alpha \cdot \left[ P^{-1}\!\left( \frac{1}{\beta}, p \right) \right]^{1/\beta},
\end{equation}
where \(P^{-1}\) is the inverse in the second argument.

Define the quantile ratio for \(0<a<b<1\):
\begin{equation}
    R(a,b) = \frac{Q(a;\alpha,\beta)}{Q(b;\alpha,\beta)}
    = \left(\frac{P^{-1}(1/\beta,a)}{P^{-1}(1/\beta,b)}\right)^{1/\beta}.
\end{equation}

\paragraph{Property 1: Scale invariance.}
The scale \(\alpha\) cancels in \(R(a,b)\), so the ratio is invariant to rescaling of a tensor.
This supports comparing tensors with different absolute update magnitudes using a common proxy.

\paragraph{Property 2: Sensitivity to distributional shape.}
Eq.~(8) shows that \(R(a,b)\) depends only on the shape parameter \(\beta\).
Intuitively, as the distribution becomes flatter (larger \(\beta\)), lower and upper quantiles become closer relative to each other, increasing \(R\); when the distribution is more peaked with heavier tails (smaller \(\beta\)), the separation between quantiles becomes larger, decreasing \(R\).
In the limiting case \(\beta\to\infty\), \(X\) approaches a uniform distribution on \([-\alpha,\alpha]\) and thus \(Y\sim\mathrm{Uniform}(0,\alpha)\), giving \(R(a,b)\to a/b\).
In practice, \(R(a,b)\) varies smoothly with \(\beta\) and provides a useful proxy for magnitude concentration.

Motivated by these properties, TADrop directly uses the quantile ratio to guide tensor-wise retention, assigning lower keep ratios to tensors with more concentrated updates and higher ratios to those with more broadly distributed magnitudes.

\begin{table*}[htpb]
\centering

\begin{adjustbox}{max width=\linewidth}
\begin{tabular}{p{5.5cm}cccccccccc}
\toprule
\rowcolor{gray!20} & Pet & Cifar-10 & Cars & MNIST & Kvasir-v2 & Food-101 & Weather & EuroSAT & Vegetables & Cats and Dogs \\  
\midrule

Individual & 92.23 & 97.88 & 85.06 & 99.22 & 94.31 & 87.94 & 98.21 & 99.11 & 100.00 & 99.03 \\
\midrule

Weight Averaging & 31.26 & 42.91 & 7.74 & 27.63 & 25.27 & 68.02 & 61.06 & 24.37 & 83.20 & 91.28 \\  
RegMean & 34.62 & 89.65 & 16.28 & 90.71 & 71.00 & 76.14 & 86.62 & 74.13 & 99.10 & 98.54 \\

Task Arithmetic & 33.24 & 59.86 & 9.34 & 30.81 & 31.05 & 73.06 & 74.56 & 31.25 & 91.97 & 93.61 \\
Ties-Merging & 12.84 & 42.82 & 5.30 & 23.21 & 21.09 & 73.22 & 72.86 & 10.98 & 92.31 & 91.88 \\
AdaMerging  & 48.34 & 87.54 & 0.42 & 81.22 & 22.76 & 75.23 & 89.13 & 44.60 & 97.97 & 96.91 \\
\midrule

EMR-Merging  & 86.50 & 96.73 & 43.50 & 97.79 & 84.07 & 83.22 & 96.97 & \textbf{97.85} & 99.96 & \textbf{99.49} \\
\hspace{10pt} \textbf{+TADrop} &\textbf{90.22} &\textbf{97.71} &\textbf{70.74} &\textbf{98.77} &\textbf{92.39} &\textbf{87.60} &\textbf{97.72} & 97.11 &\textbf{100.00} & 99.44 \\
\midrule
\rowcolor{gray!20} & DTD & Fashion & Dogs & STL-10 & Flowers & LandScape & RESISC45 & EMNIST & Intel-Images & CUB-200-2011 \\
\midrule

Individual & 71.75 & 93.26 & 89.90 & 99.07 & 98.19 & 94.00 & 98.90 & 94.67 & 94.63 & 91.88  \\
\midrule

Weight Averaging & 14.63 & 20.46 & 47.80 & 37.74 & 68.97 & 73.14 & 13.56  & 7.73 & 82.40 & 37.66 \\  
RegMean & 30.53 & 83.42 & 42.89 & 78.94  & 95.26 & 83.64 & 34.66 & 48.67 & 93.60 & 49.78 \\

Task Arithmetic & 14.73 & 37.11 & 47.65 & 39.66 & 80.68 & 79.59 & 15.50 & 11.05 & 89.03 & 41.86 \\
Ties-Merging  & 3.71 & 27.05 & 26.03 & 6.17 & 34.33 & 78.27 & 6.79 & 5.61 & 89.39 & 31.28  \\
AdaMerging & 16.68 & 76.76 & 53.09 & 68.91 & 95.69  & 81.98 & 24.83 & 18.02 & 91.02 & 48.19 \\
\midrule

EMR-Merging & 59.62 & 89.77 & 86.48 & 98.23 & 97.91 & 94.60 & \textbf{94.80} & 93.07 & 95.46 & 82.86  \\
\hspace{10pt} \textbf{+TADrop} & \textbf{61.22} & \textbf{91.64} & \textbf{89.70} & \textbf{99.04} & \textbf{97.94} & \textbf{95.00} & 93.87 & \textbf{93.73} & \textbf{95.50} & \textbf{84.98} \\
\midrule
\rowcolor{gray!20} & SVHN & Cifar-100 & Beans & SUN397 & Garbage & Animal-10N & Fruits-360 & GTSRB & MangoLeafBD & KenyanFood13 \\
\midrule
Individual & 96.21 & 89.85 & 96.87 & 87.51 & 98.83 & 92.52 & 100.00 & 95.74 & 100.00 & 82.57 \\
\midrule

Weight Averaging & 10.88 & 77.98 & 70.98 & 57.42 & 22.89 & 46.00 & 5.38  & 15.00 & 68.58 & 33.55\\  

RegMean & 66.13 & 82.59 & 92.58 & 58.58 & 65.31 & 68.74 & 19.79 & 56.96 & 98.10 & 57.11 \\

Task Arithmetic & 17.41 & 80.20 & 84.62 & 55.88 & 25.23 & 51.14 & 6.15 & 37.01 & 87.02 & 36.32 \\

Ties-Merging & 10.54 & 78.61 & 67.22 & 52.69 & 3.91 & 19.13 & 1.50 & 40.74 & 76.58 & 19.90 \\

AdaMerging & 25.70 & 84.19 & 93.38 & 64.09 & 38.54 & 66.55 & 7.94 & 59.90 & 99.13 & 48.66 \\
\midrule

EMR-Merging & 90.56 & 87.80 & 95.31 & 80.92 & \textbf{94.43} & 88.68 & 97.40 & 95.72 & 99.97 & 78.52 \\
\hspace{10pt} \textbf{+TADrop} & \textbf{91.94} & \textbf{90.44} & \textbf{96.09} & \textbf{83.21} & 94.20 & \textbf{89.62} & \textbf{99.76} & \textbf{96.63} & \textbf{100.00} & \textbf{79.26} \\
\midrule
\end{tabular}
\end{adjustbox}
\begin{adjustbox}{max width=\linewidth}
\begin{tabular}{lcccccccccc}
\vspace{-0.45cm} \\
\rowcolor{gray!20} \multicolumn{9}{c}{\textit{\textbf{Average Performance}}} \\
\midrule
\multirow{2}{*}{} & \multirow{2}{*}{Individual} & Weight & RegMean & Task Arithmetic & Ties-Merging & AdaMerging & EMR-Merging & \multirow{2}{*}{\textbf{+TADrop}} \\
&  &Averaging  & \citep{DBLP:conf/iclr/Jin0P023} & \citep{DBLP:conf/iclr/IlharcoRWSHF23} & \citep{DBLP:conf/nips/YadavTCRB23} & \citep{DBLP:conf/iclr/YangW00G0T24} & \citep{huang2024emr} & \\
\midrule

Acc& 93.98 & 42.52 & 68.14 & 48.89 & 37.53 & 60.25 & 89.59 & \textbf{91.85}  \\
\bottomrule
\end{tabular}
\end{adjustbox}
\caption{Task-specific and average performance when merging ViT-B/16 models on 30 tasks. \textbf{Bold} numbers indicate the best score within each backbone block.}
\label{tab:30_vit_full} 
\end{table*}

\begin{table*}[t]
  \centering
  \resizebox{0.98\textwidth}{!}{
  \begin{tabular}{lcccccc}
    \toprule
    & \multicolumn{3}{c}{\textbf{Chat}} & \textbf{Math} & \multicolumn{2}{c}{\textbf{Code}} \\
    \cmidrule(lr){2-4}\cmidrule(lr){5-5}\cmidrule(lr){6-7}
    \textbf{Method} &  MMLU & HellaSwag  & TruthfulQA  & GSM8K  & HumanEval  & MBPP  \\
    \midrule
    Individual & 0.4793 & 0.75 & 0.31 & 0.45 & 0.23 & 0.2 \\
    \midrule
    Ties-Merging \citep{DBLP:conf/nips/YadavTCRB23}
      & 0.4707 & 0.74 & 0.27 & 0.23 & \textbf{0.29} & \textbf{0.29} \\
    \rowcolor{gray!12}\hspace{6pt}+ TADrop(w/o)
      & 0.4768 & 0.75 & 0.27 & 0.38 & \textbf{0.29} & 0.27 \\
    \rowcolor{gray!20}\hspace{6pt}+ TADrop
      & \textbf{0.4858} & \textbf{0.76} & \textbf{0.28} & \textbf{0.41} & \textbf{0.29} & 0.26 \\
    \midrule
    EMR-Merging \citep{huang2024emr}
      & 0.4753 & 0.72 & \textbf{0.3} & 0.47 & 0.27 & 0.24 \\
    \rowcolor{gray!12}\hspace{6pt}+ TADrop(w/o)
      & 0.4860 & 0.73 & \textbf{0.3} & \textbf{0.52} & 0.27 & 0.25 \\
    \rowcolor{gray!20}\hspace{6pt}+ TADrop
      & \textbf{0.4909} & \textbf{0.75} & 0.29 & \textbf{0.52} & \textbf{0.29} & \textbf{0.27} \\
    \bottomrule
  \end{tabular}
  }
\caption{Results on chat, math, and code benchmarks for Llama-2-7B. \textbf{Bold} numbers indicate the best score within each backbone block.}
  \label{tab:llama2_results}
\end{table*}

\begin{table*}[t]
\centering
\begingroup
\scriptsize
\setlength{\tabcolsep}{3pt}
\renewcommand{\arraystretch}{0.99}
\resizebox{0.99\textwidth}{!}{
\begin{tabular}{lccccccccc}
\toprule
\textbf{Method} & \textbf{CoLA} & \textbf{SST-2} & \textbf{MRPC} & \textbf{STS-B} & \textbf{QQP} & \textbf{MNLI} & \textbf{QNLI} & \textbf{RTE} & \textbf{Avg.} \\
\midrule
Individual & 60.18 & 94.04 & 89.22 & 90.63 & 91.41 & 87.20 & 92.71 & 79.06 & 85.56 \\
\midrule
Weight Averaging & 13.96 & 64.11 & 69.36 & 31.84 & 75.36 & 42.19 & 58.70 & 55.23 & 51.34 \\
Fisher Merging~\citep{DBLP:conf/nips/MatenaR22} & 6.11 & 53.56 & 65.21 & 22.80 & 58.51 & 34.80 & 67.64 & 55.96 & 45.57 \\
RegMean~\citep{DBLP:conf/iclr/Jin0P023} & 36.67 & 90.60 & 75.74 & 62.68 & 83.55 & 70.02 & 82.35 & 58.48 & 70.01 \\
\midrule
Task Arithmetic~\citep{DBLP:conf/iclr/IlharcoRWSHF23} & 18.78 & \underline{85.89} & \textbf{79.90} & \textbf{74.03} & 83.78 & 59.08 & 69.67 & \underline{62.09} & 66.65 \\
\hspace{6pt}+ ImPart~\citep{DBLP:conf/acl/YangL0WY0C25} & 21.29 & 85.67 & \underline{79.17} & \underline{72.40} & 82.24 & 54.43 & 69.21 & 61.37 & 65.72 \\
\hspace{6pt}+ TALL-masks~\citep{DBLP:conf/icml/WangDOFF24} & 11.39 & 79.01 & 76.72 & 25.30 & 81.26 & 38.60 & 71.06 & 53.07 & 54.55 \\
\rowcolor{gray!12}\hspace{6pt}+ TADrop(w/o) & \underline{28.36} & 85.78 & \underline{79.17} & 71.58 & \textbf{84.11} & \underline{59.84} & \underline{76.26} & 61.73 & \underline{68.35} \\
\rowcolor{gray!20}\hspace{6pt}+ TADrop & \textbf{30.90} & \textbf{86.81} & 78.92 & 68.90 & \underline{83.89} & \textbf{62.59} & \textbf{81.22} & \textbf{64.98} & \textbf{69.78} \\
\midrule
TIES-Merging~\citep{DBLP:conf/nips/YadavTCRB23} & 20.48 & 84.40 & \textbf{81.13} & 58.19 & \textbf{85.70} & \underline{64.65} & 74.81 & 42.96 & 64.04 \\
\hspace{6pt}+ ImPart~\citep{DBLP:conf/acl/YangL0WY0C25} & 9.67 & 85.21 & \underline{79.90} & 69.95 & 82.09 & 52.64 & 59.47 & 52.71 & 61.45 \\
\hspace{6pt}+ TALL-masks~\citep{DBLP:conf/icml/WangDOFF24} & \textbf{31.44} & 84.52 & 78.43 & 58.21 & 83.49 & 57.41 & \underline{77.78} & \underline{60.29} & 66.45 \\
\rowcolor{gray!12}\hspace{6pt}+ TADrop(w/o) & 22.67 & \underline{85.44} & 78.43 & \textbf{73.80} & \underline{84.67} & 63.76 & 73.57 & 57.76 & \underline{67.51} \\
\rowcolor{gray!20}\hspace{6pt}+ TADrop & \underline{27.43} & \textbf{86.24} & 78.92 & \underline{72.08} & 84.48 & \textbf{66.93} & \textbf{79.68} & \textbf{61.37} & \textbf{69.64} \\
\midrule
EMR-Merging~\citep{huang2024emr} & 39.96 & 93.35 & 86.27 & 82.77 & 89.72 & 85.45 & 89.57 & 74.37 & 80.18 \\
\rowcolor{gray!12}\hspace{6pt}+ TADrop(w/o) & \textbf{62.04} & \underline{93.69} & \underline{88.48} & \underline{90.72} & \underline{90.94} & \textbf{85.96} & \underline{92.28} & \underline{75.45} & \underline{84.94} \\
\rowcolor{gray!20}\hspace{6pt}+ TADrop & \underline{61.44} & \textbf{93.92} & \textbf{89.22} & \textbf{90.76} & \textbf{91.12} & \underline{85.94} & \textbf{92.42} & \textbf{76.53} & \textbf{85.17} \\
\bottomrule
\end{tabular}
}
\endgroup
\caption{per-task performance when merging RoBERTa-base models fine-tuned on eight GLUE text classification tasks. \textbf{Bold} numbers indicate the best score within each backbone block, and \underline{underlined} numbers indicate the second-best score.}
\label{tab:roberta_glue_aligned}
\end{table*}


\begin{table*}[t!]
\centering
\resizebox{0.99\textwidth}{!}{
\begin{tabular}{lccccccccc}
\toprule
\textbf{Method} &
\textbf{SUN397} &
\textbf{Cars} &
\textbf{RESISC45} &
\textbf{EuroSAT} &
\textbf{SVHN} &
\textbf{GTSRB} &
\textbf{MNIST} &
\textbf{DTD} &
\textbf{Avg.} \\
\midrule
Task Arithmetic
& 92.66 & 81.17 & 69.27 & 36.55
& \textbf{46.15} & 42.23 & 51.23 & 71.53 & 61.35 \\
\rowcolor{gray!12}
\hspace{10pt} + TADrop(w/o)
& \textbf{94.19} & \textbf{81.51} & \textbf{71.00}
& \textbf{47.47} & 44.10 & \textbf{43.27}
& \textbf{52.05} & \textbf{72.99} & \textbf{63.32} \\
\midrule
TIES-Merging
& \textbf{94.60} & 82.30 & 69.47 & 49.91
& \textbf{45.20} & 36.52 & 58.33 & 72.63 & 63.62 \\
\rowcolor{gray!12}
\hspace{10pt} + TADrop(w/o)
& 94.52 & \textbf{82.35} & \textbf{69.76}
& \textbf{51.33} & 45.09 & \textbf{38.47}
& \textbf{58.42} & \textbf{73.63} & \textbf{64.19} \\
\midrule
KnOTS-TIES
& 92.93 & \textbf{82.87} & 71.07 & 49.87
& 54.55 & 49.24 & 71.88 & 73.99 & 68.30 \\
\rowcolor{gray!12}
\hspace{10pt} + TADrop(w/o)
& \textbf{93.53} & 81.36 & \textbf{71.43}
& \textbf{51.74} & \textbf{56.44} & \textbf{50.07}
& \textbf{74.43} & \textbf{74.27} & \textbf{69.16} \\
\bottomrule
\end{tabular}
}
\caption{Detailed LoRA merging results on eight vision classification tasks using CLIP ViT-B/32 with rank-16 adapters. \textbf{Bold} numbers indicate the best performance within each merging method group.}
\label{tab:lora_full}
\end{table*}

\end{document}